\documentclass{article}

\usepackage{amssymb}
\usepackage{latexsym}
\usepackage{float}
\usepackage{graphicx}
\usepackage{tablefootnote}
\usepackage{setspace}
\usepackage{multirow} % Span rows in tables
\usepackage{bigstrut} % Space struts in tables up and down
\usepackage{longtable} % Tables that continue onto multiple pages
\usepackage{amsmath} % \text, and other math formatting options
\usepackage{authblk}

\usepackage{url}
\usepackage[table]{xcolor}

\usepackage{hyperref}

\title{The KiTS21 Challenge: Automatic segmentation of kidneys, renal tumors, and renal cysts in corticomedullary-phase CT}

\author[1]{Nicholas Heller}
\author[2]{Fabian Isensee}
\author[2]{Dasha Trofimova}
\author[1]{Resha Tejpaul}

\author[3]{Zhongchen Zhao}
\author[3]{Huai Chen}
\author[3]{Lisheng Wang}

\author[4]{Alex Golts}
\author[4]{Daniel Khapun}
\author[4]{Daniel Shats}
\author[4]{Yoel Shoshan}
\author[4]{Flora Gilboa-Solomon}

\author[5]{Yasmeen George}

\author[6]{Xi Yang}
\author[6]{Jianpeng Zhang}
\author[7]{Jing Zhang}
\author[6,8]{Yong Xia}

\author[9]{Mengran Wu}
\author[9]{Zhiyang Liu}

\author[1]{Ed Walczak}
\author[1]{Sean McSweeney}
\author[1]{Ranveer Vasdev}
\author[1]{Chris Hornung}
\author[1]{Rafat Solaiman}
\author[1]{Jamee Schoephoerster}
\author[1]{Bailey Abernathy}
\author[1]{David Wu}
\author[1]{Safa Abdulkadir}
\author[1]{Ben Byun}
\author[1]{Justice Spriggs}
\author[1]{Griffin Struyk}
\author[1]{Alexandra Austin}
\author[1]{Ben Simpson}
\author[1]{Michael Hagstrom}
\author[10]{Sierra Virnig}
\author[11]{John French}
\author[1]{Nitin Venkatesh}
\author[12]{Sarah Chan}
\author[13]{Keenan Moore}
\author[14]{Anna Jacobsen}

\author[15]{Susan Austin}
\author[15]{Mark Austin}

\author[1]{Subodh Regmi}
\author[1]{Nikolaos Papanikolopoulos}
\author[16]{Christopher Weight}

\affil[1]{University of Minnesota, USA}
\affil[2]{German Cancer Research Center, Germany}

\affil[3]{Shanghai Jiao Tong University, China}
\affil[4]{IBM Research, Israel}
\affil[5]{Monash University, Australia}
\affil[6]{Northwestern Polytechnical University Xi'an, China}
\affil[7]{The University of Sydney, Australia}
\affil[8]{Northwestern Polytechnical University Shenzhen, China}
\affil[9]{Nankai University, China}

\affil[10]{Rocky Vista College of Osteopathic Medicine, USA}
\affil[11]{Medical Student at the University of Missouri, USA}
\affil[12]{Brigham Young University, USA}
\affil[13]{Carleton College, USA}
\affil[14]{University of Utah, USA}
\affil[15]{Mayo Clinic, USA}
\affil[16]{Cleveland Clinic, USA}
\begin{document}
\maketitle

\begin{abstract}
This paper presents the challenge report for the 2021 Kidney and Kidney Tumor Segmentation Challenge (KiTS21) held in conjunction with the 2021 international conference on Medical Image Computing and Computer Assisted Interventions (MICCAI). KiTS21 is a sequel to its first edition in 2019, and it features a variety of innovations in how the challenge was designed, in addition to a larger dataset. A novel annotation method was used to collect three separate annotations for each region of interest, and these annotations were performed in a fully transparent setting using a web-based annotation tool. Further, the KiTS21 test set was collected from an outside institution, challenging participants to develop methods that generalize well to new populations. Nonetheless, the top-performing teams achieved a significant improvement over the state of the art set in 2019, and this performance is shown to inch ever closer to human-level performance. An in-depth meta-analysis is presented describing which methods were used and how they faired on the leaderboard, as well as the characteristics of which cases generally saw good performance, and which did not. Overall KiTS21 facilitated a significant advancement in the state of the art in kidney tumor segmentation, and provides useful insights that are applicable to the field of semantic segmentation as a whole.
\end{abstract}

% \begin{keyword}
% %% MSC codes here, in the form: \MSC code \sep code
% %% or \MSC[2008] code \sep code (2000 is the default)
% % \MSC 62H35\sep 68T45\sep 62P10\sep 62M45
% %% Keywords
% \KWD Semantic Segmentation\sep Renal Cancer\sep Computed Tomography
% \end{keyword}

% \end{frontmatter}

%\linenumbers

%% main text
\section{Introduction}
\label{sec:kits21.introduction}

\subsection{Kidney Tumor Background}
With the utilization of cross-sectional imaging now as high as it's ever been, kidney tumors are now most often discovered incidentally, rather than on the basis of symptoms \cite{mcdonald2015effects,capitanio2016renal}. There is growing evidence that large numbers of renal tumors are either benign or indolent, especially when they are small and incidentally discovered, and they might therefore be best managed with surveillance rather than intervention \cite{prasad2008benign,vasudevan2006incidental,jewett2011active}. However, metastatic renal cancer remains highly lethal, so the rare instances in which a small and/or incidentally-discovered renal tumor progresses to metastatic disease are unacceptable, and their risk must be weighed against overtreatment and its associated cost and morbidity \cite{rosiello2021renal}. Some argue that renal mass biopsy has the potential to resolve this treatment decision dilemma, but others argue that its relative lack of sensitivity hinders its ability to convince physicians and patients that their disease won't progress \cite{tomaszewski2014heterogeneity}, and ultimately, its utilization remains relatively low \cite{leppert2014utilization}. There remains a significant unmet need for tools to reliably differentiate between benign/indolent renal tumors and those with metastatic potential.

\subsection{Kidney Tumor Radiomics}
Increasingly, the so-called ``radiome'' is revealing itself as a powerful quantitative predictor of clinically-meaningful outcomes in cancer \cite{gillies2016radiomics}. In renal tumors, radiomic features have shown exciting potential for predicting histologic subtype \cite{bhandari2021ct}, nuclear grade \cite{nazari2020noninvasive}, somatic tumor mutations \cite{karlo2014radiogenomics}, and even cancer-specific and overall survival \cite{han2022performance}. In surgical oncology, a number of ``nephrometry'' scores such as R.E.N.A.L. \cite{kutikov2009renal} and PADUA \cite{ficarra2009preoperative} have been developed which synthesize various manually-extracted radiomic features to produce scores which have been shown to robustly correlate with perioperative and oncologic outcomes. Of these, the R.E.N.A.L. score has recently been approximated in terms of segmentation-based radiomic features, and was shown to be noninferior to human-derived scores at predicting clinical outcomes \cite{heller2022computer}, and the others are sure to follow in short order.

\begin{figure}
    \centering
    \includegraphics[width=\columnwidth]{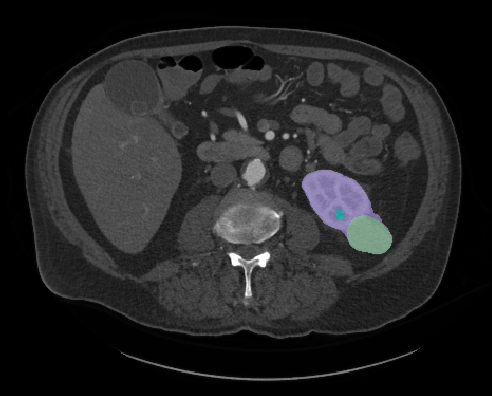}
    \caption{An example axial view of a segmentation showing all three classes represented in this challenge's dataset: `Kidney' in purple, `Tumor' in green, and `Cyst' in blue. Best viewed in color.}
    \label{fig:teaser}
\end{figure}

At the heart of most radiomics approaches is the need for the spatial delineation of which structures occupy what space in a given image. As one might imagine, manually delineating each region of interest is a very time consuming activity, and is subject to significant interobserver variability \cite{heller2018imperfect} to which, radiomics algorithms are sometimes quite sensitive \cite{liu2020stability}. There is thus significant interest in developing highly-accurate automatic methods for semantic segmentation. 

Modern deep learning approaches have achieved impressive performance on a wide variety of semantic segmentation tasks \cite{hao2020brief}, but their need for large and high-quality training datasets has hindered their development in problems such as kidney cancer for which little annotated data is publicly available. Further, the development of deep learning algorithms requires a huge number of design decisions, not only about their structure but also about procedures during training and validation. There remains little consensus in the computer vision community about which algorithms are truly optimal for a given semantic segmentation task.

\subsection{The KiTS21 Challenge}
Machine learning competitions, or ``challenges'' as they are often called, have become a mainstay in the medical image analysis research community \cite{reinke2021common}. In a challenge, a central organizing team takes responsibility for defining a clinically important problem and collecting a large labeled dataset. They then split this dataset into a public ``training set'' which is disseminated to the larger research community, and a ``test set'' which is kept secret. Teams are invited to train their favorite machine learning models on the training set, and the organizing team is responsible for measuring and ranking how well these models perform on the secret test set.

Challenges serve as an excellent model for interdisciplinary collaboration: The organizing team, which is ostensibly most interested in the domain-specific nuances of the clinical problem, benefits from top research groups from around the world turning their attention to their problem and proposing solutions. And the participants, who are ostensibly most interested in machine learning methodology, benefit from a new and high quality dataset carefully tailored to a clinically meaningful problem by domain experts. In a way, these challenges play a role analogous to that played by ``model organisms" and ``cell lines" in the biological sciences -- that is, they allow researchers to make the sort of head to head comparisons that would otherwise be impossible if everyone was working only on their own private datasets -- or their own private breed of organism, or their own private line of immortalized cells.

One of the first machine learning competitions to use this format was the \textit{Critical Assessment of protein Structure Prediction} or ``CASP'' competition which has been followed by a sequel CASP event every even-numbered year since 1994 \cite{moult2005decade}. CASP is in its 15th iteration at the time of writing, and it has served as an invaluable resource to the protein structure prediction community over the last 30 years as it has progressed through several generations of computational biochemistry \cite{kryshtafovych2021critical}, with the latest being dominated by deep learning methods such as DeepMind's "AlphaFold" \cite{jumper2021highly}. One might wonder whether DeepMind would have shown such strong interest in this important problem if it had not been so carefully and painstakingly curated into a challenge format as it was by the organizers of CASP. It is the authors' strong belief that the same applies to most challenges: they attract interest and attention to their chosen problem by individuals and research groups that otherwise might never have spent time on them.

Many excellent challenges have been organized in the medical image analysis community over the last two decades, and the semantic segmentation of cross-sectional images is one of the most popular subjects \cite{maier2018rankings}. These segmentation challenges have asked participants to segment things such as specific anatomical structures like bones and organs \cite{litjens2014evaluation, heimann2010segmentation, kavur2021chaos, sekuboyina2021verse, ma2022fast}, organs at risk in radiation therapy planning \cite{lambert2020segthor}, and, like KiTS, lesions and the organs they affect \cite{andrearczyk2021overview, bilic2022liver, bakas2018identifying}.

Of particular interest is the recent QUBIQ challenge \footnote{\url{https://qubiq21.grand-challenge.org/}} % perhaps we should make it a standard reference
which provided participants with multiple independent annotations per region of interest. Here, participants were asked to train a model not only to segment the region accurately, but also to estimate the pixelwise uncertainty in their segmentations in the hope that model uncertainty would be highest in the areas where different annotators disagreed.

The challenge described in this report, ``The 2021 Kidney Tumor Segmentation Challenge'' or ``KiTS21'' is the second challenge in the ``KiTS'' series after its 2019 iteration, ``KiTS19'' \cite{heller2021state}. KiTS19 represented the first large-scale publicly available dataset of kidney tumor Computed Tomography (CT) images with associated semantic segmentations. It attracted submissions from more than 100 teams from around the world, and saw the winning team surpass the previous state of the art in kidney tumor segmentation, while nearly matching human-level performance in segmenting the affected kidneys. KiTS21 builds upon KiTS19, but differs from it in several important ways, enumerated below:

\begin{enumerate}
    \item Data annotation was performed in public view
    \item Like QUBIQ, multiple annotations were released per ROI
    \item Renal cysts were segmented as an independent class
    \item The test set images came from a separate institution in a different geographical area
    \item Teams were required to submit a paper summarizing their method for review and approval before participating
\end{enumerate}

The following section will explain each of these new design features in detail, while also providing an in-depth description of the cohort. The remainder of this report proceeds as follows: Sec \ref{sec:materials.and.methods} describes the KiTS21 dataset and annotation process. In section \ref{sec:results}, the results of the KiTS21 challenge are discussed, including a statistical analysis of the leaderboard and the methods used by the 3 highest-performing teams. Section \ref{sec:conclusions} concludes with the a discussion of the lessons learned from organizing this challenge and possible future directions for KiTS. 

\section{Materials and Methods}
\label{sec:materials.and.methods}

\subsection{The KiTS21 Dataset}
\label{subsec:kits21.dataset}
The dataset used for KiTS21 consisted of two distinct cohorts for training and test sets collected at separate time points for different purposes. Ultimately, they were both annotated with segmentation labels in a single unified effort, but the processes to identify the patients and collect their images are described in separate sections below.

\subsubsection{Training Set Collection}
After receiving approval from the University of Minnesota institutional review board (study 1611M00821), a query was designed to identify patients from the institution's electronic medical record system who met the following criteria:

\begin{enumerate}
    \item Underwent partial or radical nephrectomy between January 1, 2011 and June 15, 2019
    \item Were diagnosed with a renal mass prior to the nephrectomy
    \item Underwent a CT scan within the 80 days prior to their nephrectomy
\end{enumerate}

This returned a collection of 962 patients. Manual chart review of each of these 962 patients was used to identify those who specifically underwent nephrectomy for fear of renal malignancy. The resulting 799 patients were reviewed in random order to identify those who had a CT scan available which showed the entirety of all kidneys and kidney tumors, and were in the corticomedullary contrast phase. After reviewing 544 cases in this way, 300 were identified for use as the training set for this study. It is important to note that these 300 cases were the same 300 that were split between the training (210) and test (90) sets for the KiTS19 challenge, \cite{heller2021state}.

\subsubsection{Test Set Collection}
% TODO: Describe this in greater depth
The test set used for the KiTS21 challenge consisted of 100 cases of patients who had been treated with partial or radical nephrectomy for fear of renal malignancy at the Cleveland Clinic. Preoperative CT images were obtained from all patients for whom they were available, and these patients were reviewed in random order until 100 patients with a scan in the corticomedullary phase were identified for use as the test set of the KiTS21 challenge.

\subsubsection{Overall Dataset Characteristics}

The characteristics of the patients comprising the training and test sets can be found in table \ref{tab:kits21.baselines}. Of note is a stark gender imbalance in which men outnumber women by roughly a 2:1 ratio. This imbalance is consistent between the training and test sets, and is, in fact, a well established phenomenon in the epidemiology of renal cell carcinoma \cite{capitanio2019epidemiology}. The median age of patients in the training set was 60 years, and the median age of patients in the test set was 63 years. The median BMI of patients in the training set was 29.82 kg/$\text{m}^2$, and the median BMI of patients in the test set was 29.7 kg/$\text{m}^2$. The median tumor diameter in the training set was 4.2 cm, and the median tumor diameter in the test set was 3.8 cm.

\begin{figure}
    \centering
    \includegraphics[width=1.0\columnwidth]{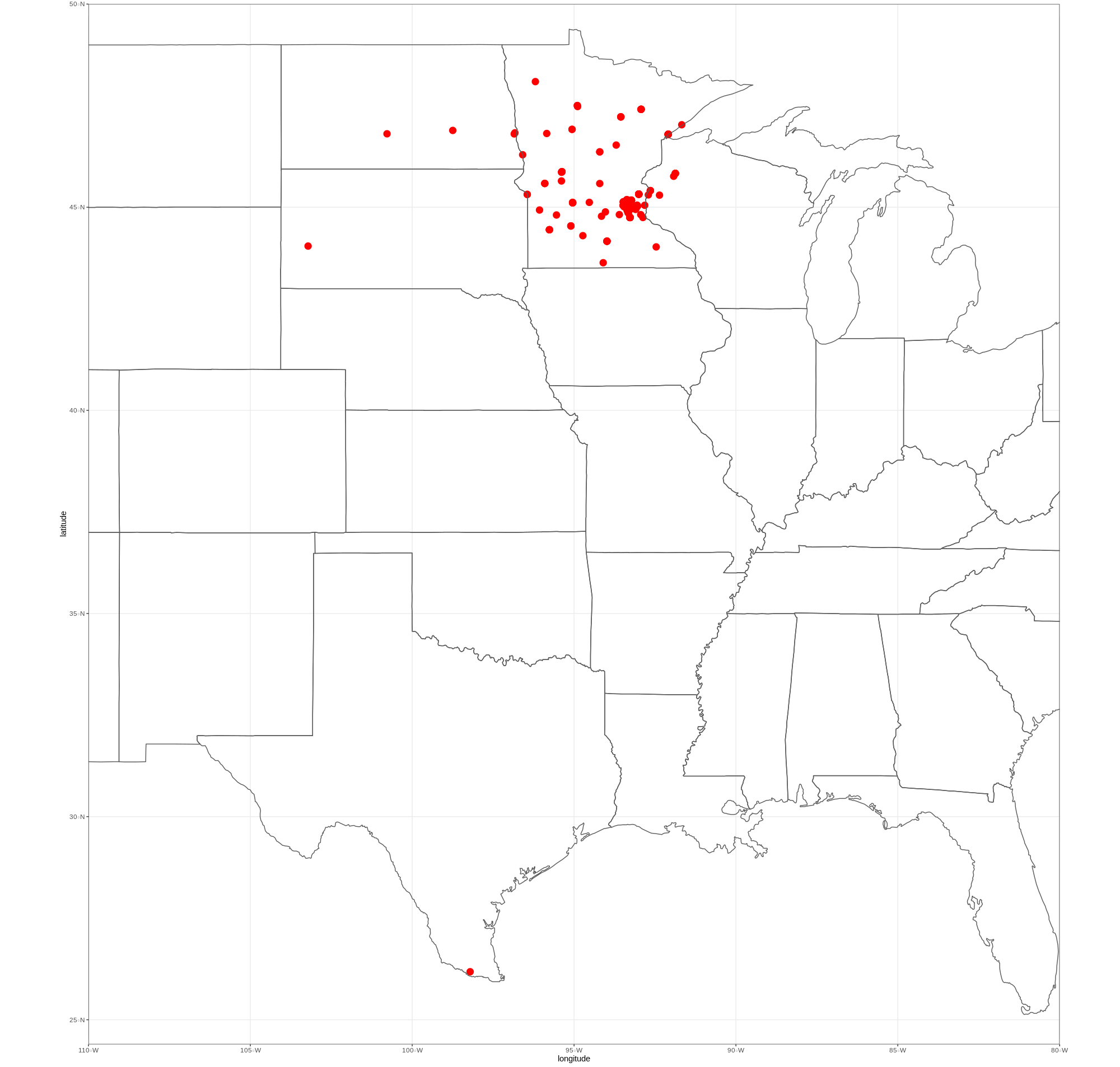}
    \caption{The geographic locations of the scanning institutions represented in the KiTS21 dataset.}
    \label{fig:kits21.geographic.locations}
\end{figure}

While the patients used for the training set were all treated at a single academic health center, the preoperative scans used in this dataset were often captured at a variety of community hospitals and clinics prior to referral. This endows the dataset with significant heterogeneity in terms of which scanners were used and with what protocol. A map depicting the geographic locations of all of the scanning institutions represented in this dataset is shown in figure \ref{fig:kits21.geographic.locations}.

\renewcommand{\arraystretch}{1}
\begin{table}[H]
{
\begin{tabular}{|p{0.3\columnwidth}|p{0.3\columnwidth}|p{0.3\columnwidth}|}
\hline

\rowcolor{gray!70} \textbf{Attribute} & \textbf{Training (N=300)} & \textbf{Testing (N=100)} \\
Age (years) & 60 (51, 68) & 63 (55, 60) \\
\rowcolor{gray!10}BMI (kg/$\text{m}^2$) & 29.82 (26.16, 35.28) & 29.7 (25.7, 33.5) \\
Tumor Diameter* (cm) & 4.2 (2.6, 6.1) & 3.8 (2.9, 5.4) \\

\rowcolor{gray!40}\multicolumn{3}{|l|}{\textbf{Gender}} \\
\hspace{0.2cm} Male & 180 (60\%) & 63 (63\%) \\
\rowcolor{gray!10}\hspace{0.2cm} Female & 120 (40\%) & 37 (37\%) \\

\rowcolor{gray!40}\multicolumn{3}{|l|}{\textbf{pT Stage}} \\
\hspace{0.2cm} pT1a & 146 (48.7\%) & 55 (55\%) \\ 
\rowcolor{gray!10}\hspace{0.2cm} pT1b & 59 (19.7\%) & 16 (16\%) \\ 
\hspace{0.2cm} pT2a & 15 (5\%) & 3 (3\%) \\ 
\rowcolor{gray!10}\hspace{0.2cm} pT2b & 5 (1.7\%) & 0 (0\%) \\
\hspace{0.2cm} pT3a & 70 (23.3\%) & 26 (26\%) \\
\rowcolor{gray!10}\hspace{0.2cm} pT4 & 5 (1.7\%) & 0 (0\%) \\

\rowcolor{gray!40}\multicolumn{3}{|l|}{\textbf{Subtype}} \\
\hspace{0.2cm} Clear Cell RCC & 203 (67.7\%) & 70 (70\%) \\ 
\rowcolor{gray!10}\hspace{0.2cm} Papillary RCC & 28 (9.3\%) & 11 (11\%) \\ 
\hspace{0.2cm} Chromophobe RCC & 27 (9\%) & 8 (8\%) \\ 
\rowcolor{gray!10}\hspace{0.2cm} Oncocytoma & 16 (5.3\%) & 4 (4\%) \\
\hspace{0.2cm} Other & 26 (8.7\%) & 7 (7\%) \\

\rowcolor{gray!40}\multicolumn{3}{|l|}{\textbf{Grade}} \\
\hspace{0.2cm} 1 & 33 (11\%) & 4 (4\%) \\ 
\rowcolor{gray!10}\hspace{0.2cm} 2 & 119 (39.7\%) & 37 (37\%) \\ 
\hspace{0.2cm} 3 & 66 (22\%) & 40 (40\%) \\ 
\rowcolor{gray!10}\hspace{0.2cm} 4 & 26 (8.7\%) & 7 (7\%) \\
\hspace{0.2cm} N/A & 56 (18.7\%) & 12 (12\%) \\
\hline
\end{tabular}
}
\caption{The baseline and tumor characteristics of patients in the KiTS19 dataset. Continuous variables are reported as: Median (Q1, Q3). *In cases where there is more than one tumor, measurement on the largest tumor is reported. The initialism RCC represents Renal Cell Carcinoma.}
\label{tab:kits21.baselines}
\end{table}

\subsection{Data Annotation Process}
Based on the prior experience and feedback collected during and after the KiTS19 challenge, KiTS21 features a unique purpose-built annotation process.

\subsubsection{Public Annotation Platform}
There has recently been some discussion about the need for greater clarity about how annotations are produced for medical image analysis research \cite{radsch2022labeling}. All too often, papers simply report which structures were segmented along with the credentials of the researcher or group of researchers who supervised and or carried out the annotations. This approach fails to capture important nuances about the annotation process such as how the regions of interest were specifically defined, what tool was used to produce the annotations themselves, and specific instructions that were given to the annotators, if any, regarding uncertainty and quality control. This information is crucial for making informed and fair comparisons regarding the performance of models on a given task.

\begin{figure}
    \centering
    \includegraphics[width=1.0\columnwidth]{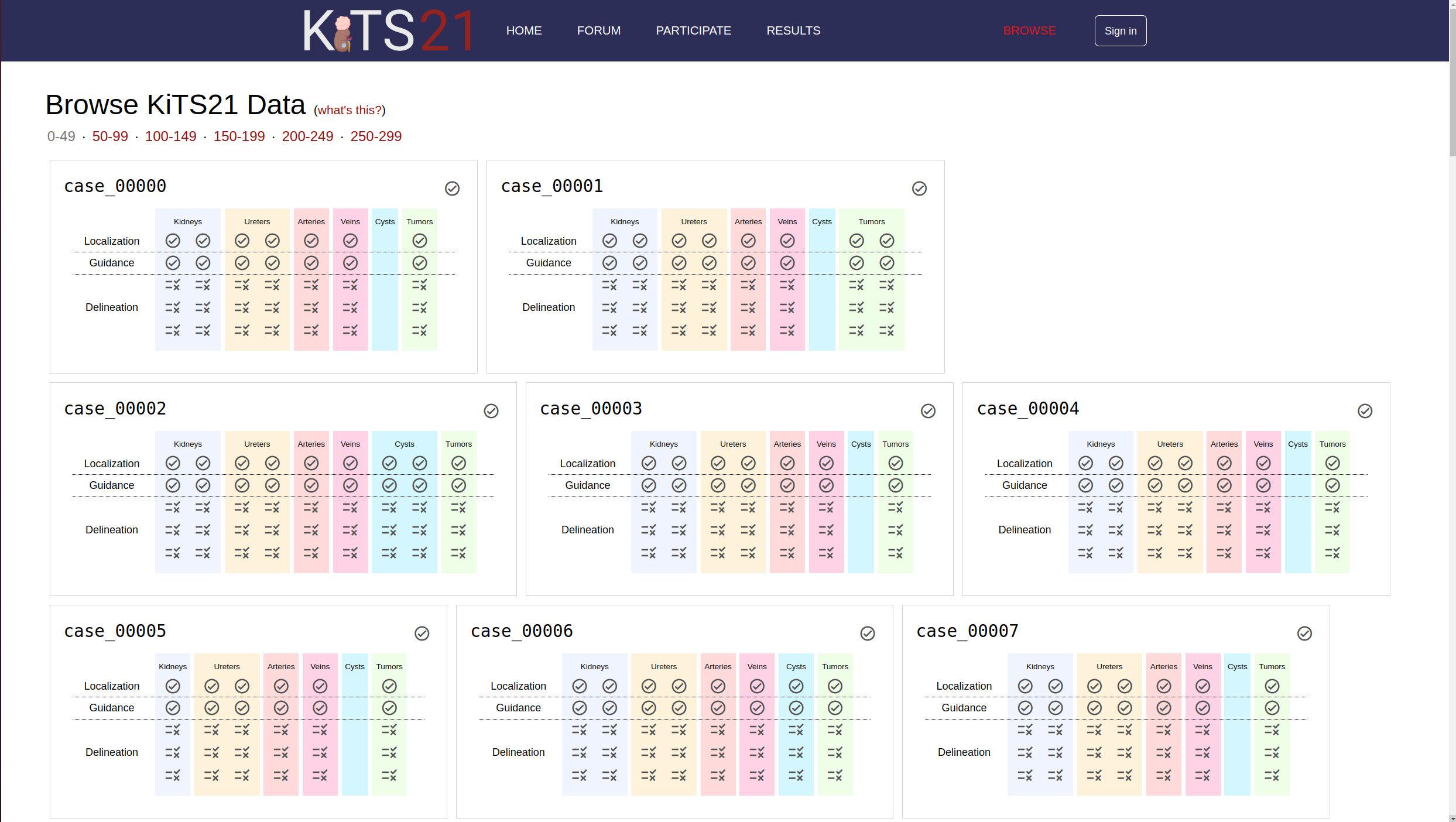}
    \caption{A screen capture showing the "Browse" page served to the public on the challenge website. Each case columns for its regions of interest with icons that represent where that case is in the annotation process.}
    \label{fig:kits21:browse}
\end{figure}

In an attempt to provide as much clarity as possible regarding the annotation process, the training set was annotated in such a way that any member of the public could view the annotation process as it took place. A website was developed\footnote{\url{https://kits21.kits-challenge.org}} which offered a dashboard showing every training case and its status in the annotation process. For an example of this, see Fig. \ref{fig:kits21:browse}. Each region of interest is denoted with a set of clickable icons that the user can use to view the annotations it represents in the same tool that the annotators used to produce them\footnote{\url{https://github.com/SenteraLLC/ulabel}}. Further still, the exact set of instructions provided to the group of annotators is documented in a webpage that is available for anyone to view\footnote{\url{https://kits21.kits-challenge.org/instructions}}.

\begin{figure}
    \centering
    \includegraphics[width=1.0\columnwidth]{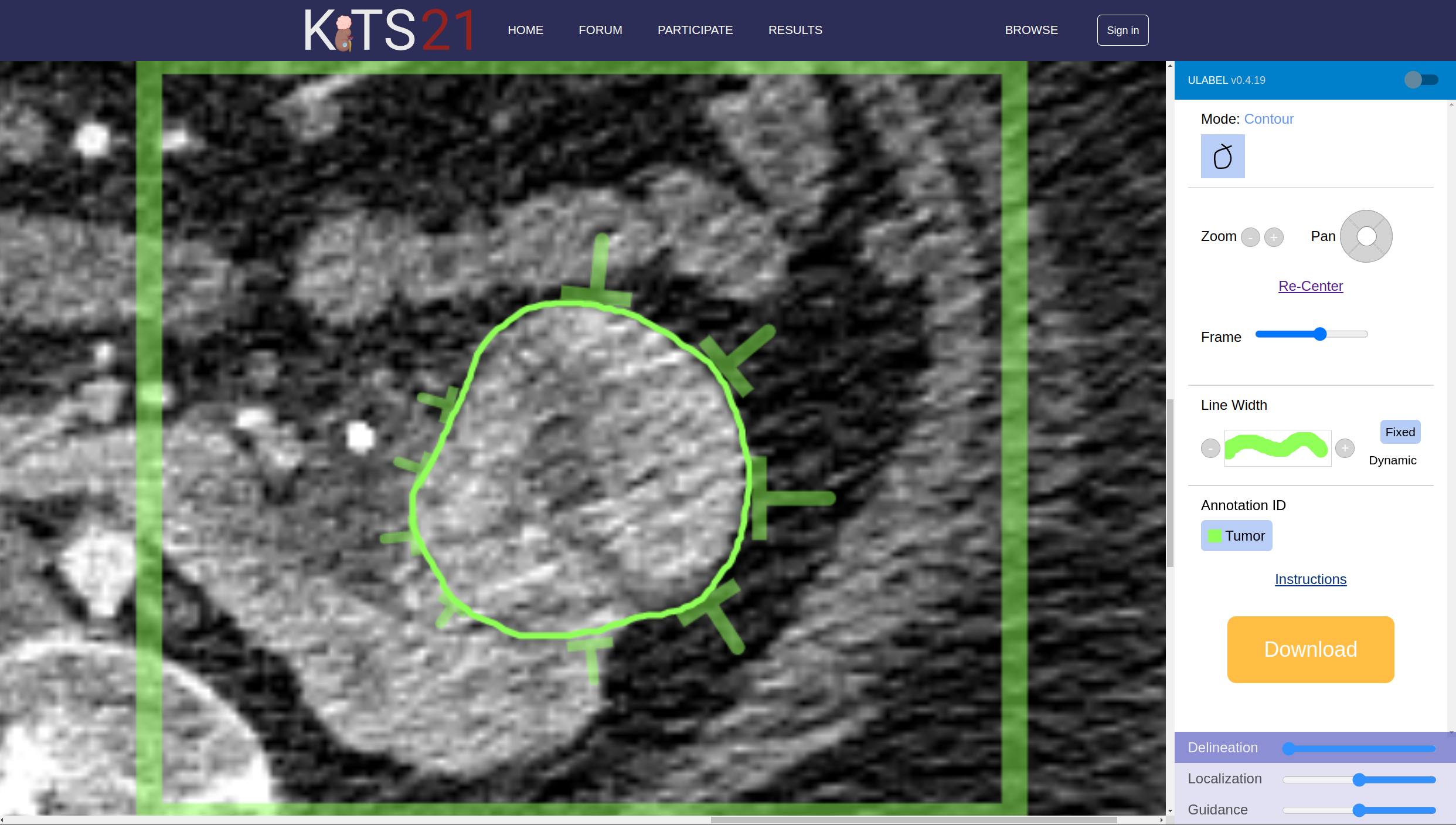}
    \caption{A screen capture showing an example of the web-based annotation tool developed for this challenge. When one of the icons shown above is clicked, it brings the user here to view and interact with those annotations.}
    \label{fig:kits21:annotate}
\end{figure}

\subsubsection{Multiple Annotations per Region of Interest}
No semantic segmentation dataset is exactly correct, nor will one ever be. For much the same reason that surgeons will excise some `margin' of healthy tissue with a tumor, a radiologist cannot always identify the exact extent of a tumor's border with 100\% certainty. This issue is further complicated by artifacts such as partial volume averaging in cross-sectional imaging.

On top of genuine uncertainty, there is a second factor which contributes to error in semantic segmentation datasets: mistakes. This is akin to `coloring outside the lines' in a coloring book. When one is asked to precisely delineate hundreds of structures with dozens of axial frames to delineate per structure, they can quickly grow tired and the extent to which their delineations match their intentions will degrade.

\begin{figure}
    \centering
    \includegraphics[width=\columnwidth]{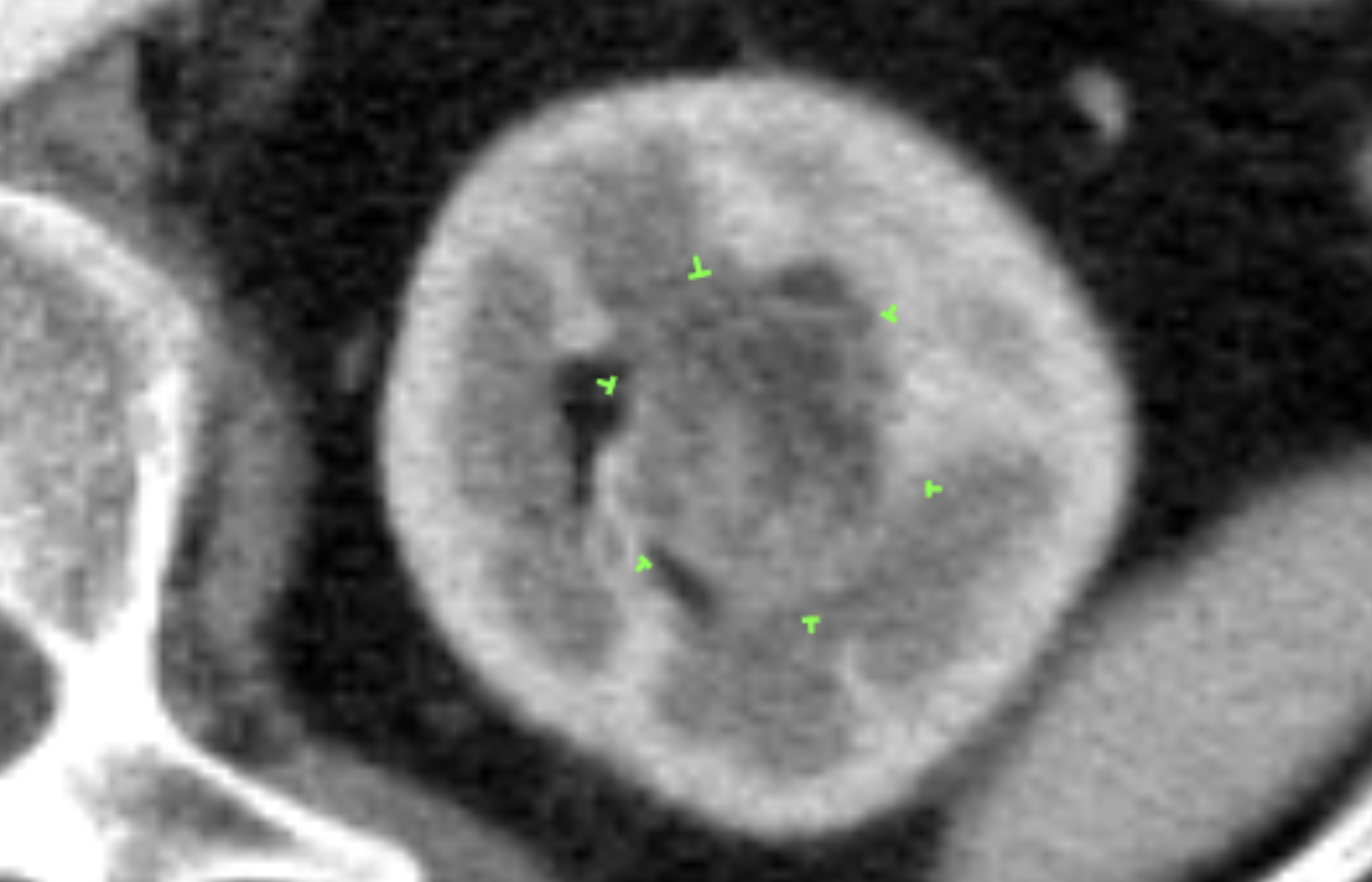}
    \caption{An example of guidance pins (green) placed by a trainee to indicate their annotation intent to guide the laypeople's delineations. Best viewed in color.}
    \label{fig:guidance.example}
\end{figure}

The KiTS21 challenge aimed to explicitly address this latter issue of delineation mistakes. To do this, the annotation process consisted of three distinct phases:

\begin{enumerate}
    \item \textbf{Localization:} A medical trainee places a 3d bounding box around the ROI
    \item \textbf{Guidance:} A medical trainee places a small number of t-shaped pins along the intended delineation path surrounding the ROI in some sample of axial slices (see Fig. \ref{fig:guidance.example})
    \item \textbf{Delineation:} A layperson (e.g., crowd worker) uses the localization and guidance to produce a delineation that matches the trainee's intentions
\end{enumerate}

In the above paradigm, localization and guidance were performed once for each region of interest, reviewed and refined by an expert if needed, and then three separate delineations were collected from laypeople. This allowed for quantifying and controlling for mistakes by collecting three independent delineations which were all guided by the same annotation intent.

\subsubsection{Changes To Intended Segmentation Classes and Dataset Size}
One might notice the presence of three additional ROI types referenced in Fig. \ref{fig:kits21:browse}: `Ureter', `Artery', and `Vein'. The original intention when planning this challenge was to segment these structures in the same way as the `Kidney', `Tumor', and `Cyst' structures that were ultimately included in the dataset. Regrettably, during the annotation process it was discovered that it would not be feasible to ensure that the segmentation labels for all six region types would be of sufficiently quality before the training set release deadline, especially with the inordinate amount of time that these complex structures took to correct and refine. The annotation team thus decided to prioritize the quality of the kidney, tumor, and cyst regions, while leaving the ureter, artery, and vein regions for future work.

\subsection{Challenge Design Decisions}

\subsubsection{Use of a Separate Institution for Test Set}
One commen criticism of single-institutional cohorts in medical image analysis is the possibility that any model trained only on that cohort might show inflated performance when validated on that cohort, as compared to its true performance on some random collection of images sampled from the true distribution of images that it is meant to be applied to. It is therefore generally recommended that data from a separate institution should be used for \textit{external validation} \cite{bluemke2020assessing}. As described in section \ref{subsec:kits21.dataset}, the KiTS21 training set is unique in that while all nephrectomy procedures took place at a single institution, patients most commonly underwent imaging at a different institution prior to being referred to the Fairview University of Minnesota Medical Center for treatment. One might therefore argue that the KiTS19 dataset already represents sufficient diversity in imaging institutions to prevent overfitting to site-specific characteristics. Nonetheless, it is much more convincing to use a properly separate cohort for the test set, and for that reason, the KiTS21 test set was built on a cohort of patients treated at a separate institution: the Cleveland Clinic.

\subsubsection{Peer Review Requirement}
One of the most important contributions a challenge can make to the research community is to elucidate which approaches work best for a particular problem. This depends on challenge participants taking the time to document their approach in a detailed publication. Unfortunately, the reports produced to accompany challenge submissions are often woefully lacking in detail and clarity, and this severely hinders a challenge's impact \cite{isensee2021nnu}.

In an attempt to prevent this, KiTS21 instituted a policy that short papers accompanying submissions to the challenge would undergo a peer review process in which they would be reviewed for clarity and completeness. After the challenge, these papers were then published as MICCAI satellite event proceedings, similar to a typical workshop. A template paper\footnote{\url{https://www.overleaf.com/read/nfbqmtkcyzdp}} was provided to teams to help guide them in what information was expected to be provided, and what a typical structure might look like. Teams were required to submit this paper at least a week before test set submissions opened, and their submissions would not be considered until the paper had been approved. Ultimately 27 out of 28 papers submitted were approved, but most had to undergo a round of revisions with repeat review.

\subsubsection{Metrics and Ranking}
In KiTS19, a simple average S\o rensen-Dice % add citation
metric was used to rank teams. S\o rensen-Dice is an attractive option because it is very widely-used and well-understood by the community. However, recent research has revealed some of its shortcomings \cite{reinke2021common2}, such as its agnosticism to how many objects were `detected' when multiple objects exist within the same image -- resulting in a case where smaller objects are given much more %should be "less", no? 
weight, when in reality, a smaller tumor on the contralateral kidney, for example, might be even more important to detect. To address this, the S\o rensen-Dice metric was supplemented with the Surface Dice metric, as described in \cite{nikolov2018deep}.

Another factor taken under consideration for KiTS21 was the natural hierarchical relationship between the target segmentation classes. Masses (i.e., tumors and cysts) are naturally part of the overall kidney region, and tumors are naturally a subset of a particular patient's collection of masses. Further, since the most difficult tasks for models to learn in this problem are (1) differentiating between tumors and cysts, and (2) determining the boundary between masses and healthy kidney, the problem was framed in terms of what we're calling `Hierarchical Evaluation Classes' (HECs) where the first class is the union of all regions, the second class is the union of the tumor and cyst regions, and the third class is the tumor alone. This prevents penalizing a model twice for mislabeling a tumor for a cyst, or part of a mass as healthy kidney, etc.

Since three independent annotations per region of interest were collected, each case consisted of $3^N$ possible composite segmentation masks, where $N$ is the number of regions of interest for that case. $N$ ranged from 3, in the simplest case (two kidneys and a tumor), to well over 10 in some cases with several cysts and tumors. At test time, a random sample of these composite segmentations for each case was used for evaluation.

Ultimately, both the S\o rensen-Dice and the Surface Dice were computed for each HEC of each randomly sampled composite segmentation in the test set. Average total scores were computed for each metric, and then a rank-then-aggregate approach was used to determine the final leaderboard rankings. In the case of a tie, the average S\o rensen-Dice score on the tumor region was used as a tiebreaker.

\subsubsection{Incentive and Prize}
Every team which had their manuscript approved for publication in the KiTS21 proceedings and made a valid submission to the challenge was invited to present their work in a short format at the KiTS21 session of the 2021 MICCAI conference. Teams that placed in the top 5 were given the option to give a longer presentation, and were also invited to participate on the challenge report as coauthors. The first place team was also awarded a cash prize of \$5,000 USD. 

\subsubsection{Changes to Intended Submission Procedure}
The testing phase of machine learning competitions generally proceeds in one of two ways. In the first way, teams are given access to the images (but not labels) of the test set for a limited period of time, during which they download the data, run inference on it with the model that they have developed, and send their predictions back to the organizers for evaluation. This was the method used by KiTS19. The second approach is to ask teams to package their model in such a way that it can be sent to the organizers and run on the test set on a private server, thereby preventing the participating team from ever having direct access to the test set images.

The second approach is generally thought to be preferable, since it eliminates all possibility that a team might `cheat' by manually intervening in the predictions made by their model to improve them unfairly. The downside to this approach, however, is that teams are limited to the computational resources made available to them by the organizers, which depending on the level of funding, might be quite limited. The sorts of large ensembles of models that often win machine learning competitions require significant time and resources to run, and might not be feasible in a challenge using the latter approach. Further still, the latter approach adds significant complexity to the tasks of both the organizers and the participants, with both parties having to build and maintain systems for these inference tasks. Containerization solutions such as Docker have made tasks like this easier, but they come with a learning curve, and not everyone in the research community has extensive experience with them.

KiTS21 originally planned to ask teams to prepare Docker containers in which to submit their models for fully private evaluation. Regrettably, soon into the submission period it became clear that this would not be practical with the resources available. More than half of the teams who submitted their containers exhausted either the time or memory constraints imposed by the cloud-based submission system that had graciously been made available by the \url{https://grand-challenge.org} platform. Ultimately the responsibility for this unfortunate debacle lies with the organizers for failing to clearly communicate the resource limitations to participants. Since teams had already invested considerable effort in solutions that exceeded the resource limitations, it was decided the most fair thing to do was to pivot to the former approach in which teams were provided with a limited 48 hour window during which to download the test imaging, run inference on their own computing hardware, and return their predictions to the organizers.

\section{Results}
\label{sec:results}

\subsection{Performance and Ranking}
% Overview in terms of number of submissions, papers, retractions, event at MICCAI etc.
Overall, 29 teams submitted papers to the KiTS21 challenge proceedings, and in-so-doing, registered their intention to submit predictions. One of these teams was not able to be reached after submitting their paper, and so was excluded. Another signed up to receive a copy of the test set images, but did not return predictions, and so was also excluded. A further team withdrew prior to receiving the test set images. This left 26 teams who submitted their predictions to the challenge. Once the results were announced, a further one team asked to have their paper retracted and their results removed from the leaderboard. All told, 25 teams were included in the final leaderboard with corresponding manuscripts. The top-5 teams were invited to give long-form oral presentations at the MICCAI KiTS21 workshop. The remaining 20 teams were given the option to give a shorter talk, and 9 of them accepted.

\begin{figure}[H]
    \centering
    \includegraphics[width=\columnwidth]{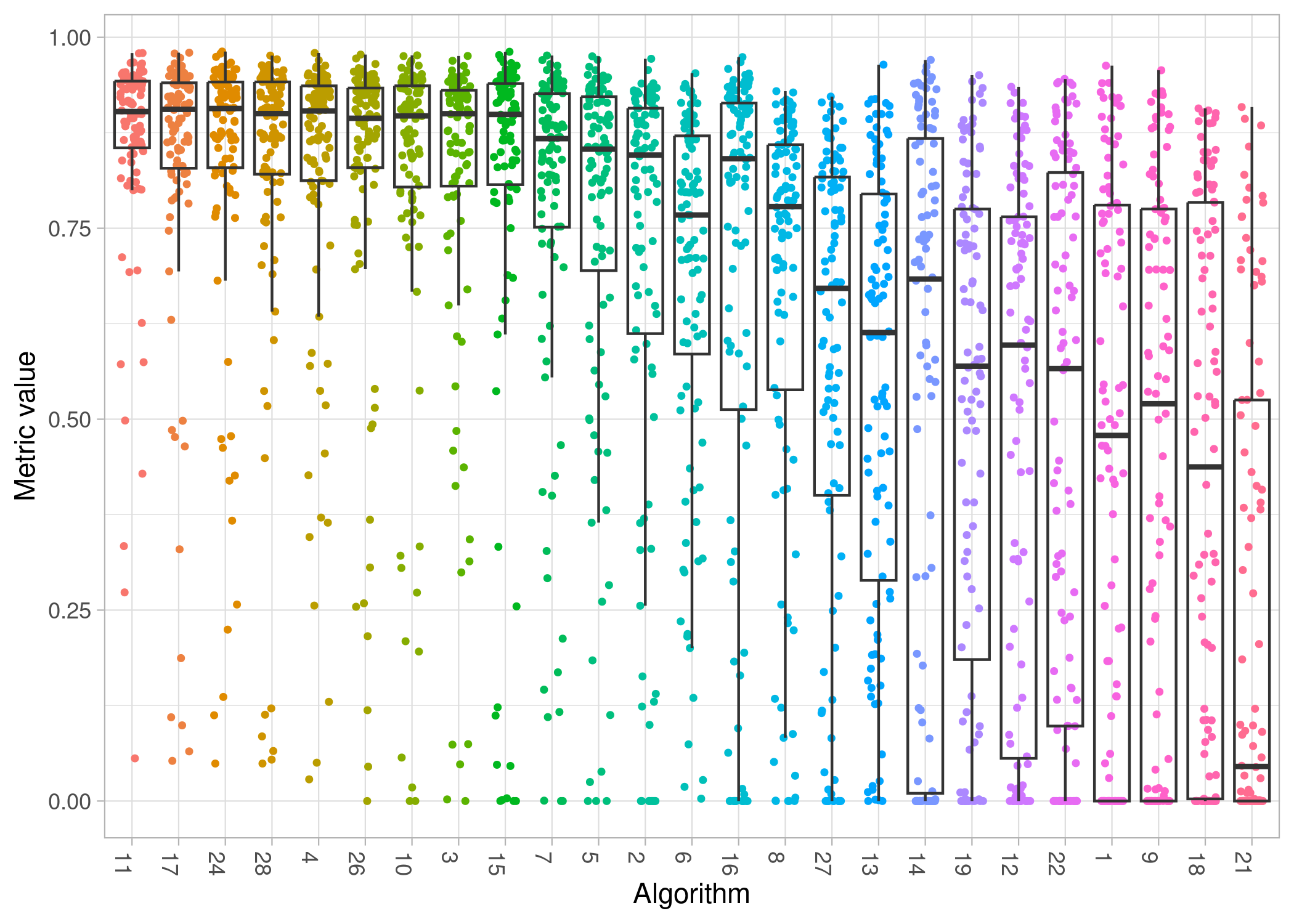}
    \caption{Box plots of tumor segmentation performance for each team on the 100 test set cases. Generated with the ChallengeR package \cite{wiesenfarth2019methods}.}
    \label{fig:kits21.box.plot}
\end{figure}

% Visualization of scores by team for kidney and tumor -- compare to human inter-rater agreement
Shown in figure \ref{fig:kits21.box.plot} is a series of box plots for each team's tumor segmentation performance on the 100 test set cases. For reference, the inter-rater agreement in tumor segmentation for this task was previously shown to be 0.88 \cite{heller2021state}, whereas the top-5 teams achieved values of 0.86, 0.83, 0.83, 0.82, and 0.81 respectively. The top-9 teams all appear to have very similar performance profiles with a tight cluster around 0.8 and then a low-density uniform distribution of scores on one or two dozen cases. Interestingly, the top 5 teams had no "complete misses" among them, with a nonzero dice score on all 100 cases. This stands in contrast to the KiTS19 leaderboard in which every team missed at least one tumor completely.

\begin{figure}[H]
    \centering
    \includegraphics[width=0.9\columnwidth]{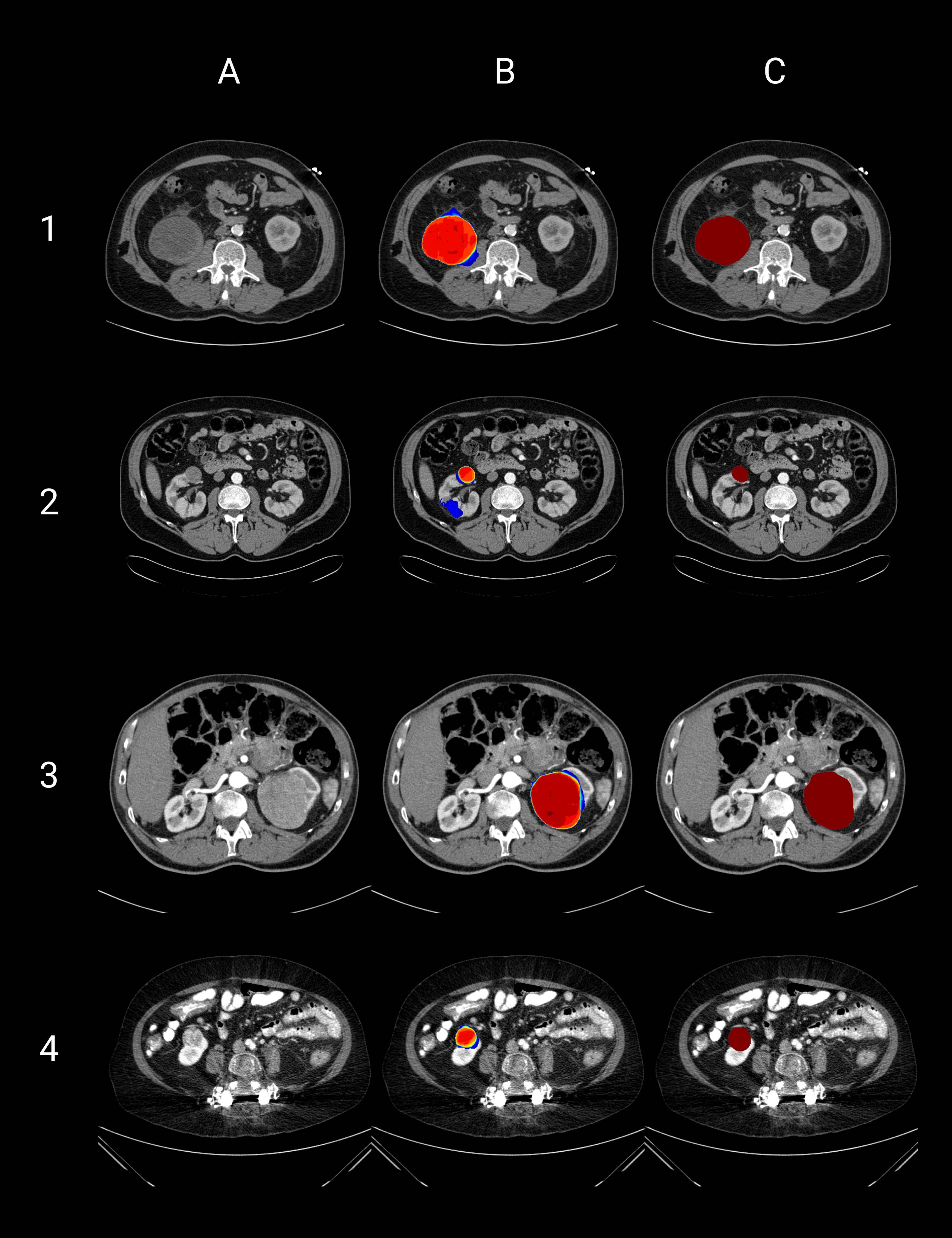}
    \caption{A heatmap showing the sum of each team's predictions on the first four cases in the test set. Column A shows the raw image, column B shows the composite heatmap for the tumor region's predictions, and column C shows the ground truth segmentation label for the tumor region.}
    \label{fig:kits21.composite.heatmap}
\end{figure}

% Visualization of composite predictions as a heatmap -- demonstrates that most difficut aspect remains identification of kidney--lesion boundary
The variety in predictions for a single case can be qualitatively examined by plotting the sum of those predictions as a heatmap. This is what is shown in figure \ref{fig:kits21.composite.heatmap}. Clearly, the vast majority of teams concentrated their predictions on the correct region of the kidney. However, there is significant variation in the exact delineation of the boundary between the lesion and the kidney. This is consistent with expert opinion that this tumor-kidney delineation is much more difficult than simply detecting the tumor itself. That said, a precise and accurate boundary delineation is very important for downstream applications of these segmentations, such as for surgical planning, where a misjudgement of the boundary could lead to the unnecessary removal of too much healthy parenchyma, leading to poor renal functional outcomes, and the removal of too little tumor could lead to positive surgical margins and a greater risk for avoidable recurrence at the primary site. 

\begin{figure}[H]
    \centering
    \includegraphics[width=\columnwidth]{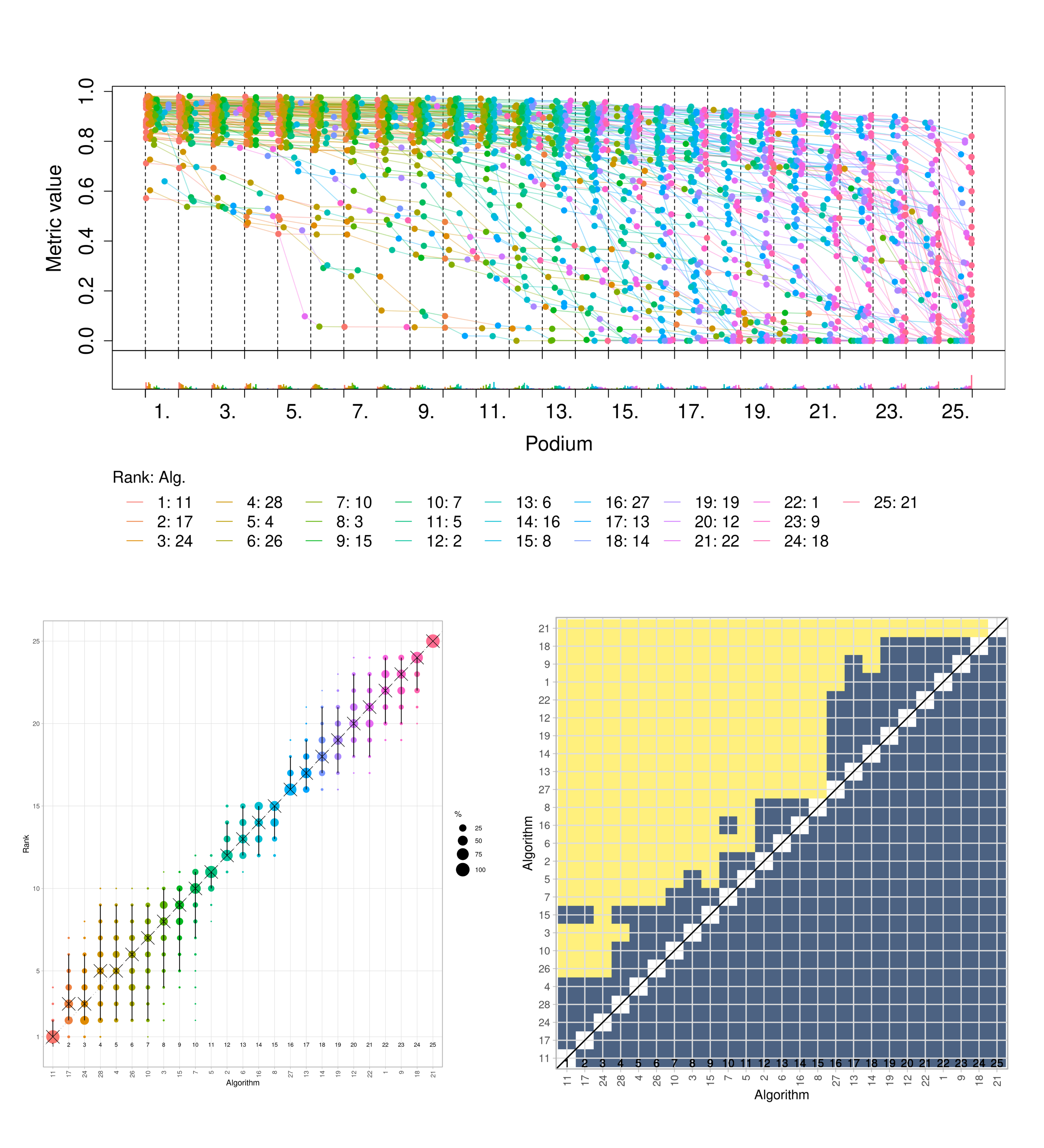}
    \caption{Three figures probing the stability of the final leaderboard ranking. At the top, a podium plot \cite{eugster2008exploratory} shows the aggregate metric values for each team for each bootstrap sample of the test set, with lines connecting each sample iteration. The bottom left shows these bootstrap samples in another way, with dot sizes depicting the frequency with which each team appeared in a given leaderboard spot. Finally, the bottom right shows a map of pairwise statistical significance after correcting for multiple hypothesis testing. Generated with the ChallengeR package \cite{wiesenfarth2019methods}.}
    \label{fig:kits21.ranking.analysis}
\end{figure}

% Statistical significance of leaderboard rankings -- tiled figure here
The final leaderboard ranking was determined with a rank-then-aggregate procedure using the respective means across HECs of the two chosen varieties of dice scores. A static final ranking is, of course, necessary in order to award prizes and to determine which teams are invited to present at the MICCAI KiTS21 workshop. However, it is important to note that the final ranking does not necessarily represent an unimpeachable truth about which teams are better than others. The test set has a finite size, and so inferential statistics must be used to make claims about differences in performance. The final results of this pairwise analysis, along with descriptive plots supporting these conclusions are shown in figure \ref{fig:kits21.ranking.analysis}. All analyses were performed with $\alpha = 0.05$ and corrections for multiple hypothesis testing were made using the Holm-Bonferroni procedure \cite{holm1979simple}.

As shown, the first-place team was not statistically superior to any of the top-5 teams at a family-wise error rate of $\alpha = 0.05$, but it was suprior to nearly every team thereafter -- with the interesting exception of the 9th place team. This lack of statistical significance is often used to criticise challenges as being ``unfair'' or ``unreliable''. In the context of the nominal cash prize awarded to the ``winning'' team, perhaps this is true. However, it can be argued that the true value in machine learning competitions is not in the pairwise testing between individual methods, but rather in the population-level meta-analyses that they enable regarding the design decisions that were made by each team. Challenges, therefore, might not be the best way to determine which method is ``best'', but they are an excellent way to determine which methods are currently ``better'' than others. It is therefore of great importance that the submissions be accompanied by detailed manuscripts describing which methods were used. KiTS21 not only requested, this, but enforced it using a peer review process. After the challenge, discrete data about each method were collected by a manual review of each paper, and a brief discussion of this data is provided in the following section.

\subsection{Methods Used}

\begin{figure}[H]
    \centering
    \includegraphics[width=\columnwidth]{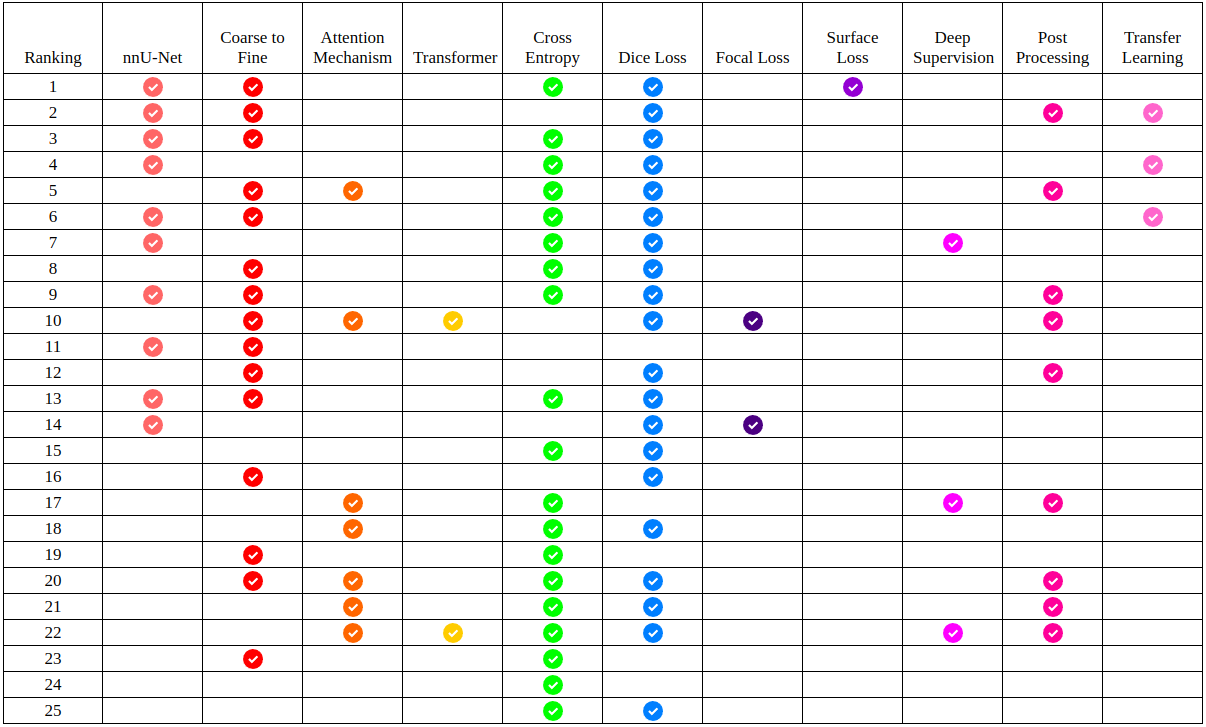}
    \caption{A tabular representation of which teams used which of a set of 11 selected methods. In the case where a single team has multiple loss functions, a weighted sum of these loss functions was used. Post processing, in this context, refers to the use of contour or connected component analysis to refine the output of the model based on some set of heuristics. Transfer learning, in this context, refers to pre-training the model on a different dataset, and then fine-tuning on the KiTS21 training set.}
    \label{fig:kits21.method.table}
\end{figure}

% Overview of methods used by teams, with citations
In the course of manually reviewing each team's manuscript, 11 specific binary data points were extracted from each paper, corresponding to 11 commonly-used methods for this problem. Which teams used which of these methods is presented in tabular form in figure \ref{fig:kits21.method.table}. As can be seen, the nnU-Net approach dominates the top half of the leaderboard, with "coarse-to-fine" frameworks and transfer learning both also being overrepresented among high-performing teams. The most popular loss function was the sum of cross entropy and dice loss, but notably, it was an nnU-Net trained with a sum of cross entropy and surface loss that ultimately won the competition. Many of the teams who chose not to use nnU-Net instead opted for architectures which incorporated some form of attention mechanism. Among these, two teams used a visual transformer network, but in general, these attention mechanisms underperformed in comparison to nnU-Net.

% Visualization of methods used by team

\subsection{Hidden Strata Analysis}
% Overview of hidden strata analysis methods
It's important to understand how the performance of these models varies on subpopulations within the greater population of patients they might be applied to. Existing health care disparities are well-documented \cite{betancourt2005cultural, chapman2013physicians, chunara2021telemedicine}, and there is a significant concern that the proliferation of predictive models in medicine will serve to exacerbate these disparities \cite{celi2022sources}.

One aspect of machine learning problems that heightens the risk of disparate performance on different populations is underrepresentation in the training set. In the case of KiTS21, the training set was drawn from patients treated at the University of Minnesota, which is a considerably more caucasian population than most other parts of the United States. Further still, given that the subject matter is kidney cancer, which has an intrinsic higher prevalence in males, females also make up a relative minority of the dataset (see table \ref{tab:kits21.baselines}). The following sections present an exploration of this issue using both hypothesis-driven and unsupervised methodologies.

% Overview of hidden strata analysis results
\subsubsection{Hypothesis-Driven Analysis}
Due to substantial underrepresentation of non-white and female patients in the KiTS21 training set, a natural hypothesis is that segmentation performance could vary on the basis of race and/or sex. A multivariate linear regression analysis was performed using both race and gender as predictors with two additional covariates in order to determine whether these variables independently associate with the mean model performance of the top-5 teams. The results, shown in table \ref{tab:kits21.hidden_strata} reveal that non-white patients do in fact see signficantly worse performance compared to white patients. Surprisingly, however, women actually see significantly \textit{better} performance than men. This speaks to the inherent unpredictability of hidden strata analysis, and how training set characteristics alone are not sufficient to predict how a model will perform on a given subpopulation.

% Visualization of hidden strata analysis results
\begin{table}[H]
    {
    \begin{tabular}{|p{0.35\columnwidth}|p{0.18\columnwidth}|p{0.30\columnwidth}|}
    \hline
    \rowcolor{gray!70} \textbf{Variable} & \textbf{Coefficient} & \textbf{P-value} \\
    Tumor Size (cm)* & $0.0164$ & $0.048$ \\
    \rowcolor{gray!10}Clear Cell Subtype & $-0.0138$ & $0.683$ \\
    Female Gender* & $0.0781$ & $0.022$ \\
    \rowcolor{gray!10}Non-Caucasian Race* & $-0.119$ & $0.005$ \\
    Intercept* & $0.6706$ & $0.000$ \\
    \hline\end{tabular}
    }
    \caption{Multivariate regression against average tumor dice across the top 5 teams for each case. Statistically significant p values at $\alpha = 0.05$ are marked with asterisks.}
    \label{tab:kits21.hidden_strata}
\end{table}

Interestingly, for the two teams who submitted transformer networks, the results look quite different (table \ref{tab:kits21.hidden_strata_tformer}). In fact, both gender and race fall out of significance, whereas tumor size appears to play a much larger role. This could suggest that transformer networks are more robust to hidden strata than the nnU-Net dominated top-5 submissions. However, it should be noted that the top-5 teams still outperform the transformer networks even on those subpopulations where it performs worst.

\begin{table}[H]
    {
    \begin{tabular}{|p{0.35\columnwidth}|p{0.18\columnwidth}|p{0.30\columnwidth}|}
    \hline
    \rowcolor{gray!70} \textbf{Variable} & \textbf{Coefficient} & \textbf{P-value} \\
    Tumor Size (cm)* & $0.0653$ & $0.000$ \\
    \rowcolor{gray!10}Clear Cell Subtype & $-0.0569$ & $0.149$ \\
    Female Gender & $0.0268$ & $0.493$ \\
    \rowcolor{gray!10}Non-Caucasian Race & $-0.0061$ & $0.900$ \\
    Intercept* & $0.2984$ & $0.000$ \\
    \hline\end{tabular}
    }
    \caption{Multivariate regression against average tumor dice across the teams that submitted a method based on transformer networks. Statistically significant p values at $\alpha = 0.05$ are marked with asterisks.}
    \label{tab:kits21.hidden_strata_tformer}
\end{table}

\subsubsection{Unsupervised Analysis}
Deep neural networks are very effective at extracting high-level semantic information from high-dimensional data such as images. Researchers often exploit this property in conjunction with a variety of nonparametric dimensional reduction techniques \cite{van2008visualizing, mcinnes2018umap} in order to visualize the clusters of input samples which appear to be represented by the network in similar ways. This has proven to be a useful tool for discovering structure in high-dimensional datasets. By a similar argument, the performance of a variety of deep neural networks on a given dataset can also be conceived as a high-dimensional representation of high-level semantic information about each instance. The same techniques can therefore be applied to visualize the clusters of patients which appear to have similar signatures in terms of semantic segmentation performance.

\begin{figure}[H]
    \centering
    \includegraphics[width=\columnwidth]{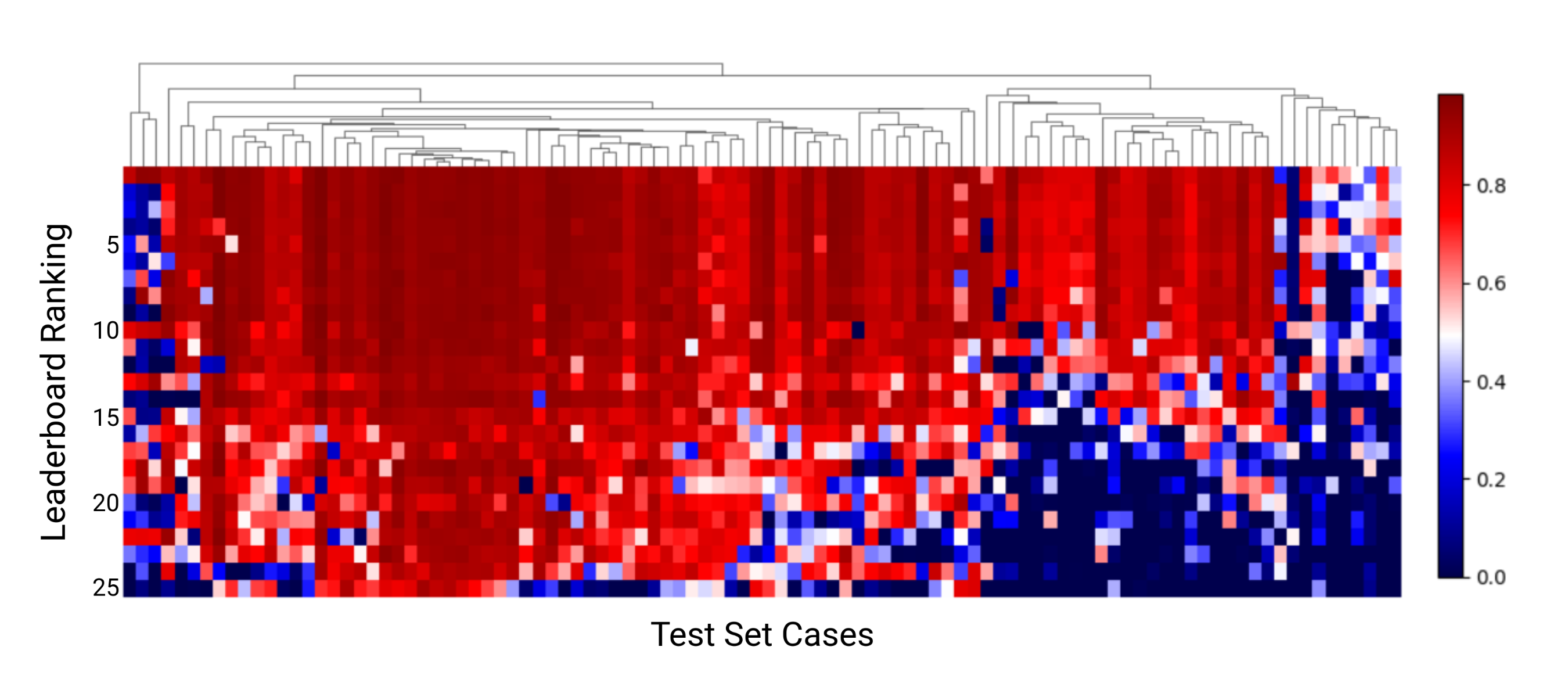}
    \caption{Hierarchical clustering of test set cases using the tumor dice scores from each team as a feature vector.}
    \label{fig:kits21.hierarchical.clustering}
\end{figure}

Figure \ref{fig:kits21.hierarchical.clustering} shows the results of hierarchical clustering performed on the set of test set cases using the performance metrics from each team as a feature vector. This reveals certain interesting clusters of cases, such as that on the far right in which virtually every team performed poorly. This stands in contrast to that cluster near the middle on which nearly every team performed well. Perhaps the most distinctive is the cluster of three cases on the far left on which nearly every teams performed poorly, except for a select few teams who performed well. Interestingly, the teams that performed well on these cases were not necessarily the teams who ranked near the top of the leaderboard.

\subsection{Methods Used by Top 3 Teams}

The three subsections that follow are brief overviews of the methods used by the three highest-performing teams who submitted to the challenge.

\subsubsection{First Place: A Coarse-to-Fine Framework for the 2021 Kidney and Kidney Tumor Segmentation Challenge}

This submission \cite{zhao2022coarse} was made by Zhongchen Zhao, Huai Chen, and Lisheng Wang from the Shanghai Jiao Tong University, China.

\paragraph{Data use and preprocessing}
This submission made use of the KiTS21 dataset alone, and used random weight initializations for their network. The images were preprocessed by resampling to an isotropic 0.78125 mm pixel size using third order spline interpolation.

\paragraph{Architectures}

\begin{figure}[H]
	\centering 
	\setlength\tabcolsep{1pt}
	\label{fig:first_place_overview}
        \includegraphics[width=0.95\linewidth]{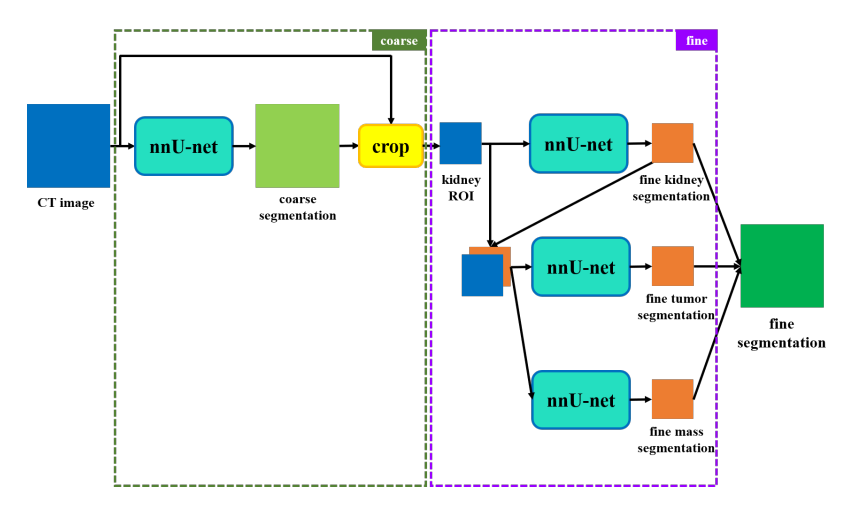}
	\caption{A graphical overview showing the coarse-to-fine paradigm used by the top-performing team.}
\end{figure}

As shown in figure \ref{fig:first_place_overview}, a coarse-to-fine approach was used which first roughly segmented the entire kidney region. This coarse kidney segmentation was used to generate a cropped region around each kidney, which was fed to a finer kidney segmentation network. The result of that network was fed to two additional networks as inputs, along with the cropped image, to produce fine tumor and "mass" segmentations, where masses refer to the union of the tumor and cyst regions. The predictions of each of these networks were aggregated to produce a final composite prediction. Each of the four networks in use were trained using the nnU-Net framework \cite{isensee2021nnu}.

\paragraph{Training}

The networks were originally trained using a sum of the dice loss and the cross entropy loss, but once these objective functions plateaued, a novel ``surface loss'' was used for further fine-tuning. The aim of this was to optimize the predictions in such a way that was in line with the surface dice metric, which was used in addition to the volumetric dice for the final leaderboard ranking. This surface loss term is defined below:

\[
L_s = \sum_{p_{pred}\in FP\cup FN} \left(\underset{p_{gt}\in S_{gt}}{\text{min}}\left|\left|p_{pred}-p_{gt}\right|\right|_2\right) * \frac{1}{C}
\]

Where $S_{gt}$ is the surface of the ground truth, $FP$ and $FN$ are the sets of false-positive and false-negative points, respectively, and $C$ is a constant.

\paragraph{Postprocessing}

Once raw predictions by the network had been made, connected component analysis was used to clean up extraneous predictions. Size thresholds of 20,000 voxels, 200 voxels, and 50 voxels were used to eliminate kidney, tumor, and cyst predictions that were too small to be realistic. Cysts and tumors that were not touching regions that were predicted to be kidney were also excluded.

\paragraph{Results}

This method achieved the $1^{st}$ place rank on the leaderboard of the KiTS21 challenge with an average volumetric dice score of 0.908 and an average surface dice score of 0.826. Of note, this method achieved a volumetric dice score for the kidney region of 0.86, which is quite close to the previously reported interobserver agreement for the KiTS19 challenge of 0.88 \cite{heller2021state}.

\subsubsection{Second Place: An Ensemble of 3D U-Net Based Models for Segmentation of Kidney and Masses in CT Scans}
This submission \cite{golts2022ensemble} was made by Alex Golts, Daniel Khapun, Daniel Shats, Yoel Shoshan and Flora Gilboa-Solomon of IBM Research - Israel.

\paragraph{Data use and preprocessing}
This submission did not make direct use of any data other than the official training set. One of the models in the final ensemble used by this method was initialized with weights of a model pretrained on the publicly available Liver Tumor Segmentation (LiTS) dataset \cite{bilic2022liver}. Other models in the ensemble were initialized randomly. This method used both low and high resolution architectures. For the former, the data was resampled to a common spacing of $1.99 \times 1.99 \times 1.99$ mm, and for the latter, $0.78 \times 0.78 \times 0.78$ mm. The labeled annotations maps used during training were sampled randomly per slice from different plausible annotations per region based on the existing multiple human annotators. This was done to improve the robustness of the trained models and make them better suited to the official KiTS21 evaluation protocol. 

\paragraph{Architectures}
A single-stage 3D U-Net and a two-stage 3D U-Net Cascade architecture were used by this method, as implemented in the nnU-Net framework \cite{isensee2021nnu}. The latter consists of first applying a 3D U-Net on low resolution data, and then using the low-res segmentation results to augment the input to another 3D U-Net applied to high resolution data. This serves the purpose of increasing the spatial contextual information that the network sees, while keeping a feasible input patch size with regards to available GPU memory. The final model ensemble used by this submission consists of three single-stage 3D U-Net models and one two-stage cascaded model.  

\paragraph{Training}
Patches of size $128 \times 128 \times 128$ were sampled and input to the network. All models were trained with a combination of Dice and cross-entropy losses \cite{isensee2021nnu}. One of the models in the final ensemble was additionally trained with a regularized loss which encouraged smoothness in the network predictions. Training was done for $\sim$250,000 iterations of Stochastic Gradient Descent. Training a single-stage 3D U-Net model took $\sim$48 hours on a single Tesla V100 GPU. All models were trained on 5 cross-validation splits with 240 cases used for training and the remaining 60 for validation.
\paragraph{Postprocessing}
Custom postprocessing was applied to the segmentation results, removing rarely occuring implausible findings: tumor and cyst findings positioned outside of kidney findings, and small kidney fragment findings surrounded by another class.
Fig.~\ref{fig:golts_et_al_examples} shows prediction examples for two slices, with and without the proposed postprocessing.
\begin{table}[H]
\label{fig:golts_et_al_examples}
\begin{tabular}{ccccc}
\textbf{\footnotesize{input}} & \textbf{\footnotesize{without}} & \textbf{\footnotesize{with}} & \textbf{\footnotesize{ground truth}} &\\
& \textbf{\footnotesize{postprocessing}} & \textbf{\footnotesize{postprocessing}} & & \\
\includegraphics[width=0.20\textwidth]{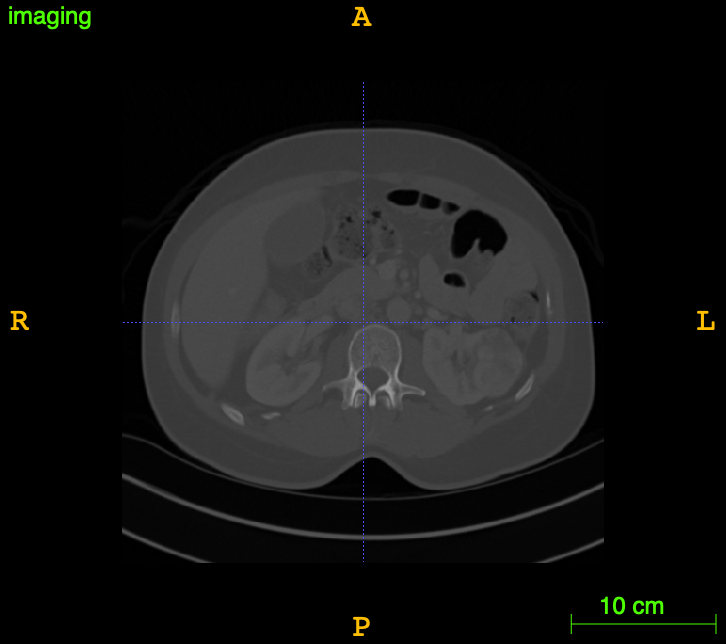} & \includegraphics[width=0.20\textwidth]{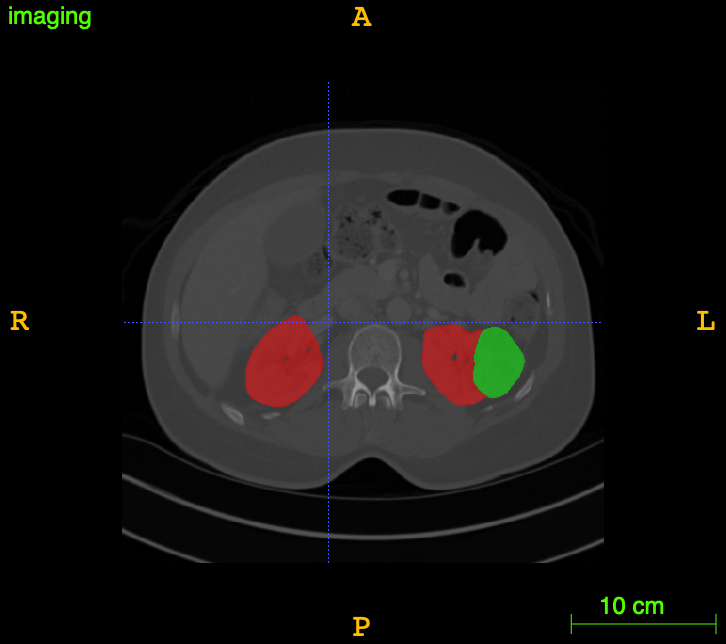} & \includegraphics[width=0.20\textwidth]{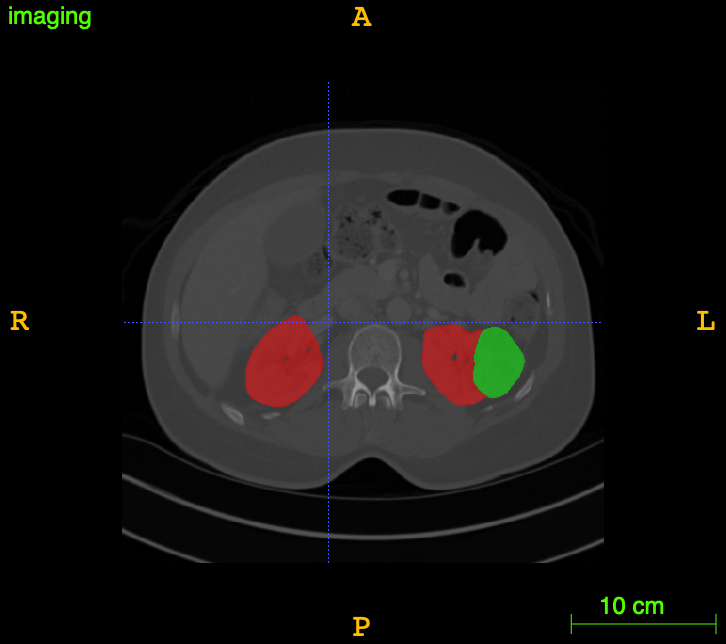} & \includegraphics[width=0.20\textwidth]{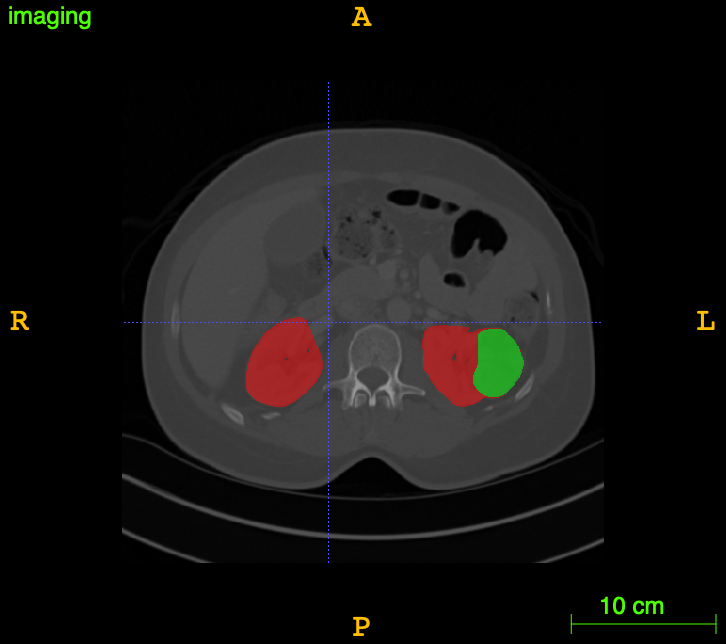} &  \\
\includegraphics[width=0.20\textwidth]{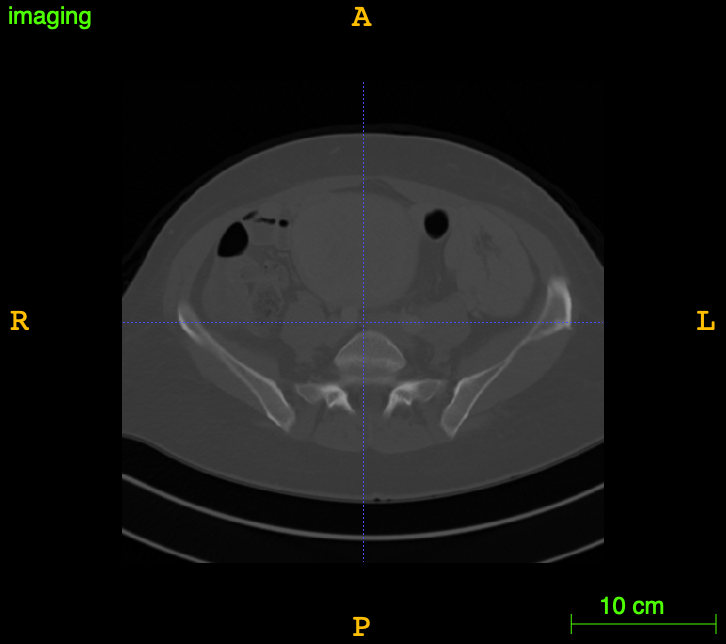} & \includegraphics[width=0.20\textwidth]{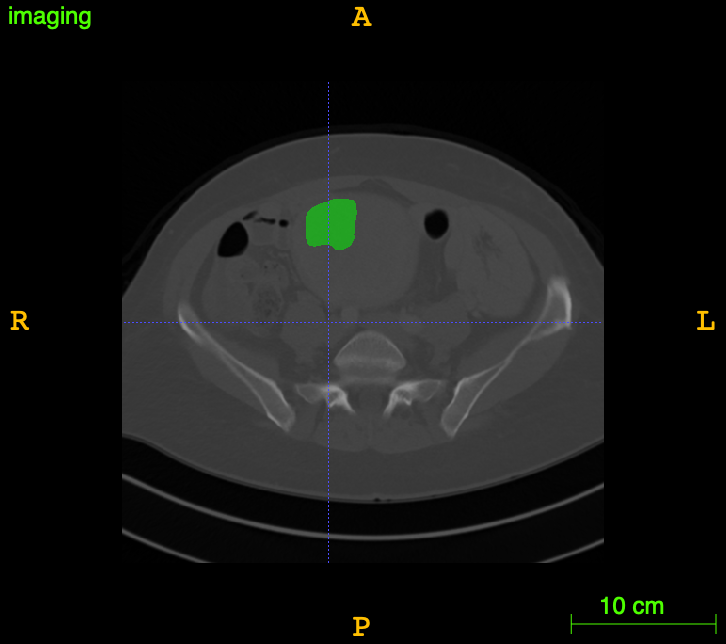} & \includegraphics[width=0.20\textwidth]{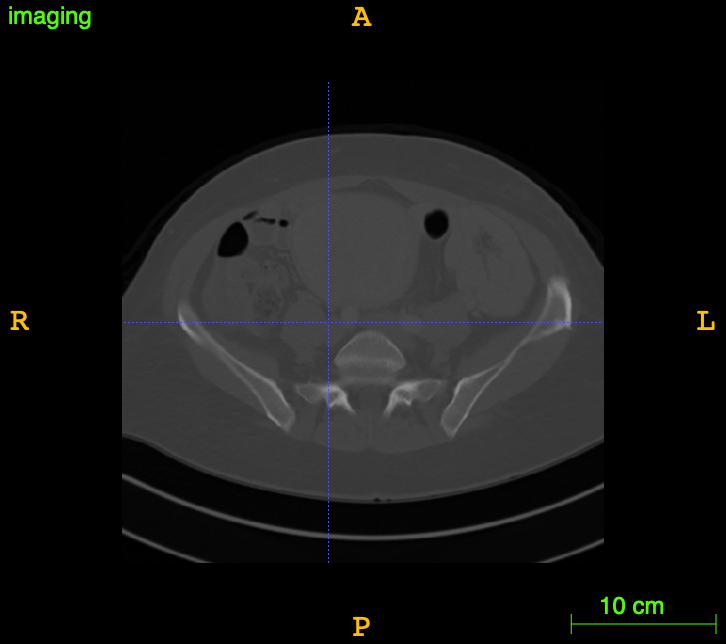} & \includegraphics[width=0.20\textwidth]{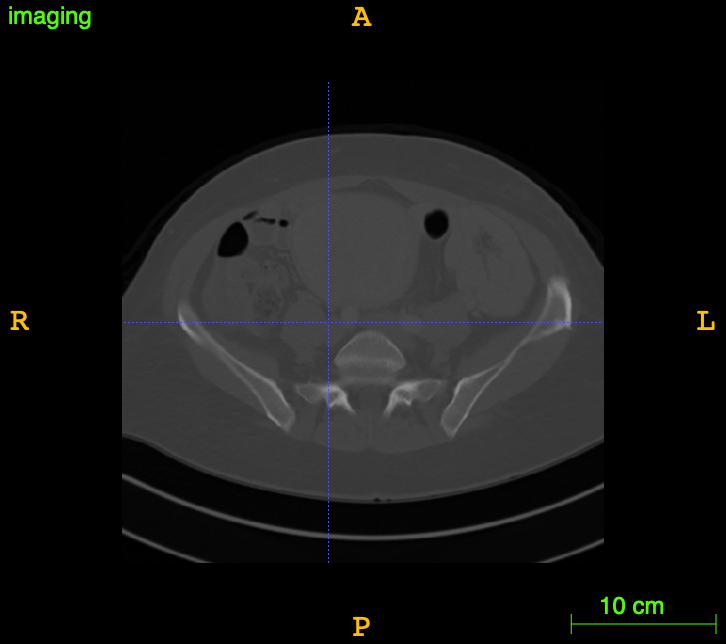} & 
\end{tabular}
\caption{Example predictions. Top row: a slice which contains kidney (red) and tumor (green) findings. Bottom row: a slice which exhibited a false tumor prediction, successfully removed by the postprocessing.}
\end{table}

\paragraph{Results}
The final submission was an ensemble of four models as follows: (1) 3D U-Net trained with an added regularized loss, (2) 3D U-Net trained with a different random seed for the training label generation process, (3) 3D U-Net and (4) 3D U-Net cascade, both trained with weights initialized from a model trained on the LiTS dataset. This submission scored 0.896 mean Dice and 0.816 mean Surface Dice, resulting in second place. For a more detailed description of this submission, see \cite{golts2022ensemble}.

\subsubsection{Third Place: 
A Coarse-to-Fine 3D U-Net Network for Semantic Segmentation of Kidney CT Scans}

This submission \cite{george2022coarse} was made by Yasmeen George from the Monash University, Australia.

\paragraph{Network architecture.} The proposed coarse-to-fine cascaded U-Net approach is based on 3D U-Net architecture and has two stages. In the first stage, a 3D U-Net model is trained on downsampled images to roughly delineate kidney region. In the second stage, a 3D U-Net model is trained to have more detailed segmentation of the three classes (kidney, tumor, cyst) using the full resolution images guided by the first stage segmentation maps. The 3D U-Net architecture had an encoder and a decoder path each with five resolution steps. The encoder part was performed using strided convolutions starting with 30 feature maps then doubling up each level to a maximum of 320. The decoder part was based on transposed convolutions. Each layer consists 3D convolution with $3\times3\times3$ kernel and strides of 1 in each dimension, leaky ReLU activations, and instance normalization. For more details please refer to our paper \cite{george2022coarse}.

\paragraph{Data preprocessing.} The CT intensities (HU) were transformed by subtracting mean and dividing by standard deviation. In the first stage, each CT scan was resampled using third order spline interpolation to a spacing of $1.99\times1.99\times1.99$ mm resulting in median volume dimensions of $207\times201\times201$ voxels. While in the second stage, a spacing of $0.78\times0.78\times0.78$ mm was used with median volume dimensions of $528\times512\times512$ voxels. Data augmentation methods including random rotations, gamma transformation, and random cropping were used during training. 

\paragraph{Training and validation.} The proposed models were implemented using nnU-Net framework \cite{isensee2021nnu} with Python 3.6 and PyTorch framework on NVIDIA Tesla V100 GPUs. Majority aggregation ground truth was used for training and validation. All models were trained from scratch using 5-fold cross-validation with a patch size of $128\times128\times128$ that was randomly sampled from the input resampled volumes. The models were trained using stochastic gradient descent (SGD) optimizer for 1000 epochs using a batch of size 2 with 250 batches per epoch. The training objective was to minimize the sum of cross-entropy and dice loss.

\begin{figure}[H]
	\centering 
	\setlength\tabcolsep{1pt}
	\label{fig:ct_pred_results}
	\begin{tabular}{c}
		\includegraphics[width=0.95\linewidth]{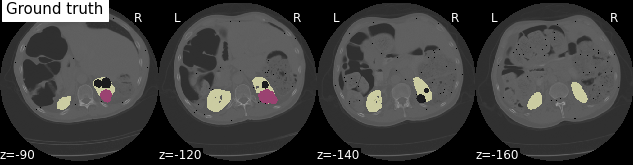} \\ 
		\includegraphics[width=0.95\linewidth]{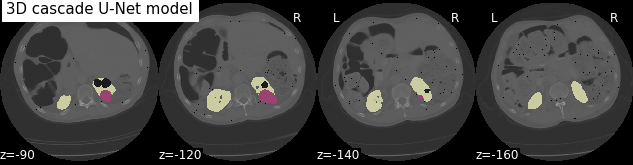}
	\end{tabular}
	\caption{Segmentation results for kidney, tumor and cyst using fine-to-coarse cascaded U-Net model}
\end{figure}

\paragraph{Results.} The model achieved the $3^{rd}$ place in the leaderboard of KiTS21 challenge with a mean sampled average dice score of 0.8944 and a mean sampled average surface dice score of  0.8140 using a test set of 100 CT scans. The proposed approach scored 0.9748,	0.8813,	0.8710 average dice for kidney, tumor and cyst using 3D cascade U-Net model. Figure \ref{fig:ct_pred_results}.visualizes the segmentation results for the trained model. 

% \subsubsection{Fourth Place: }
% % TODO
% \cite{yang2022transfer}

% \subsubsection{Fifth Place: }
% % TODO
% \cite{wu2022less}

\section{Conclusions}
\label{sec:conclusions}

This paper presented the results of the 2021 Kidney Tumor Segmentation Challenge (KiTS21). The challenge featured many innovations in terms of challenge design, including a novel annotation scheme to produce multiple volumetric annotations per ROI and a fully transparant web-based annotation process. The challenge attracted 25 full submissions from teams around the world, and the top-performing team surpassed the prior state of the art performance set with the predecessor KiTS19 challenge, despite the use of a test set from an entirely different institution and geographic area. A meta-analysis of the methods used by participating teams showed the continued popularity and dominant performance of the nnU-Net framework, although significant interest by teams in developing transformer or other attention-based methods was also observed. A hidden strata analysis was presented, which revealed that the top-performing teams were not necessarily the ones who had the most uniform performance on subpopulations within the test set.

The continued goals with KiTS are to continue to expand and augment the quality of the dataset to facilitate even better performance, while also continuing to challenge participants with additional hetereogeneity and complexity, such that the community can continue to move towards a more realistic real-world setting for this problem. The upcoming KiTS23 edition achieves this by incorporating the venous contrast phase in addition to the corticomedullary phase which both KiTS19 and KiTS21 were based on. As organizers, our hope is that the community will continue to find KiTS to be a useful resource for advancing the state of the art in kidney tumor segmentation.

\section*{Acknowledgements}

Research reported in this publication was supported in part by the National Cancer Institute of the National Institutes of Health under Award Number R01CA225435. The content is solely the responsibility of the authors and does not necessarily represent the official views of the National Institutes of Health.

Additional support for research activities including developing the annotation procedure, performing image annotations, and analyzing the submission data was provided by The Intutive Foundation, Cisco, and by research scholarships from the Climb 4 Kidney Cancer Foundation. The monetary prize for the winning team was graciously sponsored by Histosonics, Inc.

Finally, we would like to thank the urology departments at the University of Minnesota and Cleveland Clinic for graciously allowing us to use their collections of patient data for this purpose.

%% Bibliography
%%Harvard
% \bibliographystyle{model2-names.bst}\biboptions{authoryear}
\bibliographystyle{plain}
\bibliography{main}

%\section*{Supplementary Material}

%Supplementary material that may be helpful in the review process should
%be prepared and provided as a separate electronic file. That file can
%then be transformed into PDF format and submitted along with the
%manuscript and graphic files to the appropriate editorial office.

\end{document}